\documentclass[runningheads]{llncs}
\usepackage{graphicx}
\usepackage{subcaption}
\usepackage{booktabs}
\usepackage{xcolor}
\usepackage{hyperref}
\usepackage{comment}

\begin{document}
%

\title{SciBERT-based Semantification of Bioassays in the Open Research Knowledge Graph
}

\titlerunning{SciBERT-based Semantification of Bioassays}
%
\author{Marco Anteghini\inst{1,2}\orcidID{0000-0003-2794-3853} \and Jennifer D'Souza\inst{3}\orcidID{0000-0002-6616-9509}\and Vitor A.P. Martins dos Santos\inst{1,2}\orcidID{0000-0002-2352-9017} \and S\"oren Auer\inst{3}\orcidID{0000-0002-0698-2864}}
\authorrunning{Anteghini et al.}

\institute{Lifeglimmer GmbH, Markelstr. 38, 12163 Berlin, Germany
\and
Wageningen University \& Research, Laboratory of Systems \& Synthetic Biology, Stippeneng 4, 6708 WE, Wageningen, The Netherlands \email{\{anteghini,vds\}@lifeglimmer.com} \\ \and
TIB Leibniz Information Centre for Science and Technology, Hannover, Germany  \\
\email{\{jennifer.dsouza,soeren.auer\}@tib.eu}}
\maketitle              
\begin{abstract}

As a novel contribution to the problem of \textit{semantifying biological assays}, in this paper, we propose a neural-network-based approach to automatically semantify, thereby structure, unstructured bioassay text descriptions. Experimental evaluations, to this end, show promise as the neural-based semantification significantly outperforms a naive frequency-based baseline approach. Specifically, the neural method attains 72\% $F1$ versus 47\% $F1$ from the frequency-based method.

The work in this paper aligns with the present cutting-edge trend of the scholarly knowledge digitalization impetus which aim to convert the long-standing document-based format of scholarly content into knowledge graphs (KG). To this end, our selected data domain of bioassays are a prime candidate for structuring into KGs.


\end{abstract}
\section{Introduction}



Biological assays are defined as standard biochemical test procedures used to determine the concentration or potency of a stimulus (physical, chemical, or biological) by its effect on living cells or tissues~\cite{hoskins1962uses,irwin1953statistical}. 

In the context of the current Covid-19 pandemic, bioassays are critical, for example, for vaccine development. They reveal the functional and biologically relevant immunological responses that correlate with vaccine efficacy. However, massive volumes of bioassays are being produced and researchers are inundated with this information. Apart from their sheer quantity, bioassay diversity presents enormous challenges to organizing, standardizing, and integrating the data with the goal to maximize their scientific and ultimately their public health impact as the screening results are carried forward into drug development programs. 

Against this broad societal application setting, we present a solution as a step in the easier knowledge acquisition of bioassays for researchers: \textit{the neural-based automated structuring of unstructured, non-standardized bioassays based on the standardized BioAssay Ontology (BAO)~\cite{bao}}. Bioassays, until their recent semantification in an expert-annotated dataset~\cite{clark2014fast,schurer2011bioassay,vempati2012formalization} based on the BAO, were published in the form of unstructured text. Integrating their semantified counterpart in a KG facilitates their advanced computational processing. E.g., bioassays can be easily compared across their key properties, viz. Target, Perturbagen, Participants, and Detection Technology, captured as KG nodes and links. 
Nonetheless, the fine-grained semantification of bioassays as a manual task is a costly and time-intensive endeavor. Their automated semantification not only alleviates the costly manual task, but potentially makes it possible to rapidly semantify this data in large volumes. Herein, we present our novel SciBERT-based~\cite{Beltagy2019SciBERTPC} neural BAO~\cite{bao} bioassay semantification system. 

\section{Method}

For automated bioassay semantification, we carry out the supervised machine learning of semantic statements (i.e., subject-predicate-object triples) based on the BioAssay Ontology (BAO)~\cite{bao} for a given unstructured bioassay description. The code for our method is publicly available at: \footnotesize{\url{https://github.com/MarcoAnteghini/SciBERT-bioassays_ORKG}}.

\subsection{Dataset} 
Our dataset for learning comprises an expert manually annotated collection of 983 semantified bioasssays~\cite{schurer2011bioassay,vempati2012formalization}. In the data, each assay has between 5 and 92 semantic statements at an average of 53. To better reflect the data, we show example annotations in Table~\ref{tab1} for a selected bioassay.

\begin{table}
\small
\begin{tabular}{l} \\
\textsc{has assay format} $\rightarrow$ \textsc{biochemical format} \\
\textsc{has assay format} $\rightarrow$ \textsc{protein format} \\
\textsc{has assay format} $\rightarrow$ \textsc{single protein format} \\
\textsc{assay measurement type} $\rightarrow$ \textsc{endpoint assay}
\end{tabular}
\caption{\small{Four example semantic statement annotations (from 50 total) for PubChem Assay ID 346. Note, these statements are triples with subject ``bioassay.''}}
\label{tab1}
\end{table}

\subsection{Problem Formulation} 
The dataset can be formalized as follows. Let $b$ be a bioassay from the assays dataset $B$. Each $b_i$ is annotated with an annotation sequence $as_i$ such that $as_i \in S$, where $S$ is a set of all possible semantic statements seen in the training dataset. Specifically, $as_i = \{s_1, s_2, s_3, ..., s_k\}$, such that $s_x$ is a semantic statement $\in S$; $as_i$ has $k$ different statements. In general, annotation sequences are of varying lengths. 
The dataset we use has $|S| = $ 1756 unique statements  (after filtering for non-informative ones).

In the supervised task, the input data instance corresponds to a pair $(b,s; c)$ where $c \in \{true, false\}$ is the classification label. Thus, specifically, our semantification problem is formulated as a \textit{binary classification task}. $(b,s)$ is $true$ if $s \in$ $b$'s annotation sequence ($as$), else $false$. Where $false$ instances are formed by pairing $b$ with any other label not in the annotation sequence $as$ of $b$. As an aggregate, the semantification of each bioassay is a multi-label, multi-class classification problem which we have broken up into binary classification decisions.

Intuitively, our task formulation is meaningful because it emulates the way the human expert annotates the data. Basically, the expert, from their memory of all semantic statements $S$, simply assigns $s$ to a given $b$ if they deem it as $true$; irrelevant statements are not considered, thus implicitly deemed $false$.

\subsection{SciBERT-based Machine Learning}

Our machine learning system is the state-of-the-art, bidirectional transformer-based SciBERT~\cite{Beltagy2019SciBERTPC}, pre-trained on millions of scientific articles. 
In each data instance $(b,s; c)$, the classifier input representation for the pair `$b,s$' is the standard SciBERT format, treating them as sentence pairs separated by the special [SEP] token; the special classification token ([CLS]) remains the first token of every instance. Its final hidden state is used as the aggregate sequence representation for classification tasks fed into a linear classification layer.

\section{Experiments}

\subsection{Experimental Setup}
For robust evaluations, we perform 3-fold cross validation (2:1 train-test split). In each fold experiment, training data contains roughly 655 bioassays and the remaining 328 bioassays are used for testing, where the test assays are unique across the folds. Standard precision ($P$), recall ($R$), and f-score ($F1$) metrics are used. We refer the reader to the SciBERT paper~\cite{Beltagy2019SciBERTPC} for hyperparameter details. Finally, we have an additional parameter: \textit{$false$ instances per bioassay}. They are varied between 100 to 300, in increments of 10, to obtain an optimal model.

\begin{table}
\begin{small}
\parbox{.45\linewidth}{
\centering
\begin{small}
    \begin{tabular}{p{1.2cm}p{1.2cm}p{1.2cm}p{1cm}}\toprule
       $false$ labels & $P$ &  $R$ & $F1$  \\\midrule
       100 & 0.517 & 0.968 & 0.674 \\
       ... & ... & ... & ... \\
       160 & 0.549 & 0.931 & 0.688 \\
       \textbf{170} & \textbf{0.600} & \textbf{0.939} & \textbf{0.729} \\
       180 & 0.573 & 0.945 & 0.711 \\
       ... & ... & ... & ... \\
       300 & 0.471 & 0.674 & 0.551 \\\bottomrule
       \\
      \end{tabular}
\end{small}
    \caption{\footnotesize{Bioassay semantification results from five training optimization with different $false$ classification instances (full table in appendix)}}
      \label{tab:false}    
}
\hfill
\parbox{.45\linewidth}{
\centering
\begin{small}
    \begin{tabular}{p{1.8cm}p{1.2cm}p{1.2cm}p{0.9cm}}\toprule
       test set & $P$ &  $R$ & $F1$  \\\midrule
       1st fold & 0.600 & 0.939 & 0.729 \\
       2nd fold & 0.573 & 0.956 & 0.713 \\
       3rd fold & 0.589 & 0.936 & 0.719 \\
       \textbf{$Avg.$} & \textbf{0.588} & \textbf{0.944} & \textbf{0.720}
       \\\bottomrule
       \\
      \end{tabular}
\end{small}
    \caption{\footnotesize{Automatic bioassay semantification results from 3-fold cross validation with the optimal number of $false$ classification labels (170). 
    }}
      \label{tab:cv}    
}
\end{small}
\end{table}

\subsection{Results and Discussion}
\begin{figure*}[!tb]
\begin{small}
\begin{subfigure}{\textwidth}
  \centering
  \includegraphics[width=\textwidth]{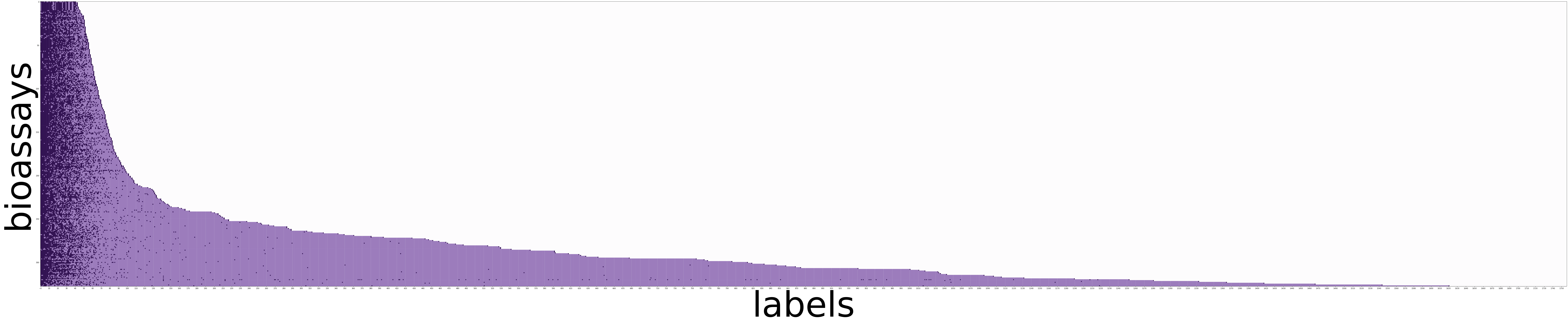}
  \caption{SciBERT classifier}
  \label{fig:sub-first}
\end{subfigure}
\begin{subfigure}{\textwidth}
  \centering
  \includegraphics[width=\textwidth]{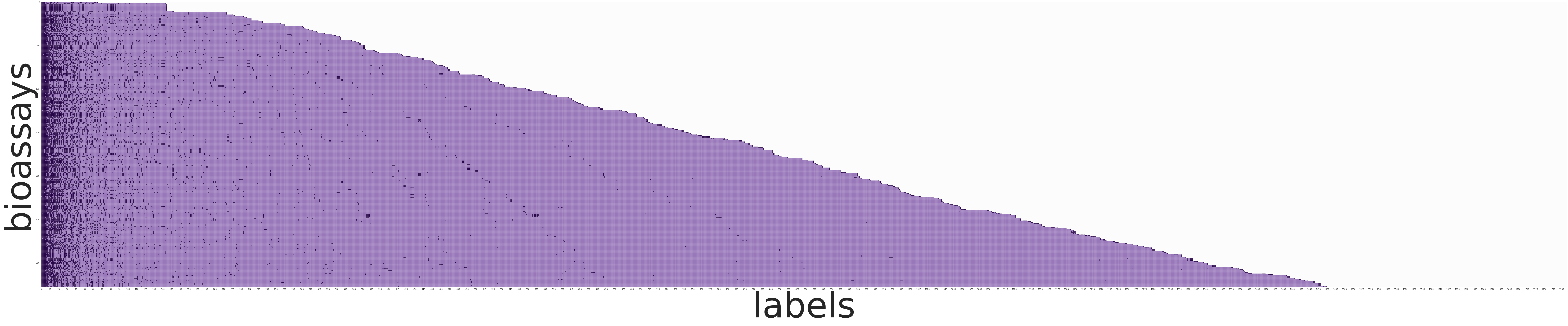}
  \caption{Frequency-based classifier}
  \label{fig:sub-second}
\end{subfigure} 
\caption{\footnotesize{Hit-and-miss Plots for semantifying bioassays by SciBERT vs. a naive frequency-based approach. Black dot is a hit; purple dot is a miss. For each assay, after all the $true$ statements are predicted, the remaining dots are white.}}
\label{fig:fig}
\end{small}
\end{figure*}

Our results are depicted in Tables \ref{tab:false} and \ref{tab:cv}. And we examine the \textit{\textbf{RQ}: can advanced neural technologies be leveraged to automatically semantify bioassays?} We find that the cumulative obtainable $F1$ by the SciBERT classifier out-of-the-box is 0.72 (bold in Table~\ref{tab:cv})---significantly higher than 0.47 from a naive frequency-based semantification approach. Furthermore, the difference of the neural approach from the frequency method is clearly evident in the hit-and-miss illustration in Fig~\ref{fig:fig}. The top thin neck of the curve in Fig 1(a) indicates that the neural approach, for most bioassays, had faster $true$ semantic statement hits among its top-scoring predictions. Thus, answering \textbf{RQ}, neural technologies can indeed perform reliable semantification of bioassays. They are also practically efficient, since, given the 1756 unique statements considered as labels, each test assay is semantified at a rate of 4 seconds.

\section{Conclusion}

The discovery of cures during pandemics such as Covid-19 can be greatly expedited if scientists are given intelligent information access tools, and our work toward automatically semantifying bioassays are a step in this direction. We refer the reader to the Appendix for an illustrated use case of semantified bioassays data in next-generation digital libraries.

\bibliographystyle{splncs04}
\bibliography{document}

\clearpage
\appendix

\section{Unique statements (labels) distribution}
\begin{figure}
    \centering
    \includegraphics[width=\textwidth]{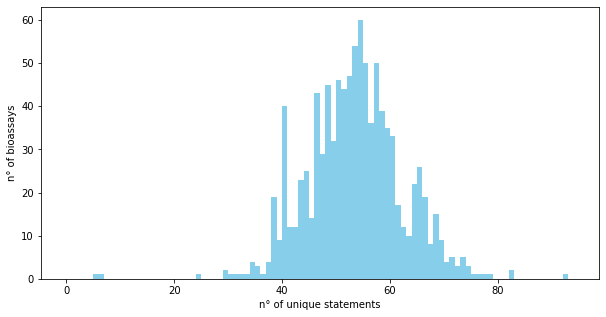}
    \caption{Unique statements distribution}
    \label{fig:labels}
\end{figure}

Each bioassays present on average 53 labels. The distribution is visible in Figure \ref{fig:labels}

\section{Snapshot of Semantified Bioassay in the Open Research Knowledge Graph}
\begin{figure*}[!h]
  \centering
  \includegraphics[width=\textwidth]{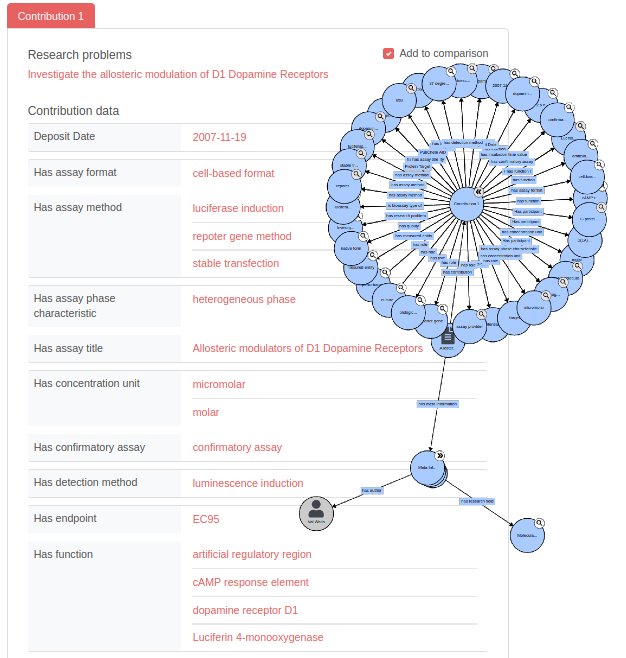}
\caption{An ORKG representation of a semantified Bioassay with an overlayed graph view of the assay. Accessible at: https://www.orkg.org/orkg/paper/R48146/R48147}
\label{fig:fig3}
\end{figure*}

\begin{figure*}[!b]
  \centering
  \includegraphics[width=\textwidth]{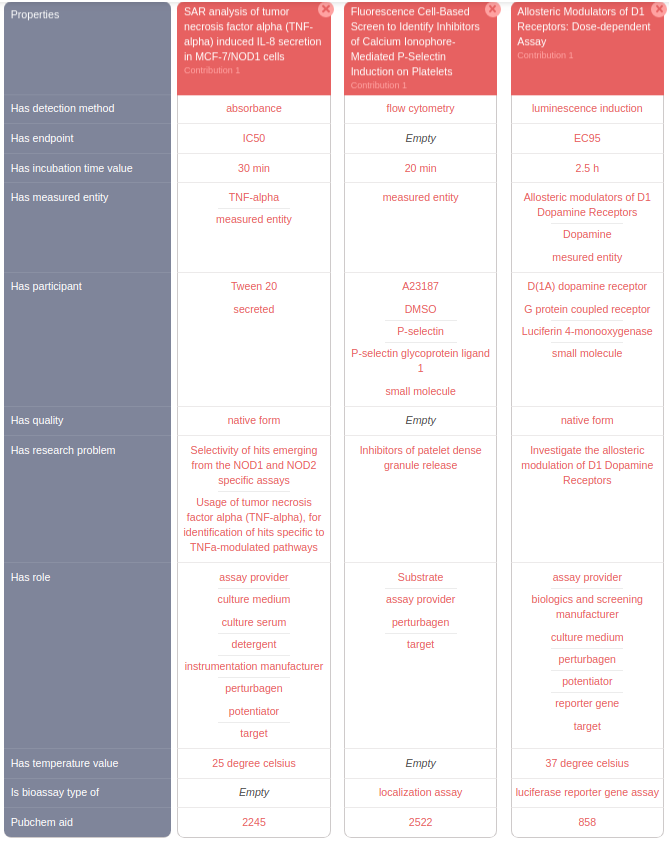}
\caption{Automatically generated comparisons of semantified bioassays in the ORKG digital library (DL). Full graph \footnotesize{\url{https://www.orkg.org/orkg/comparison?contributions=R48195,R48179,R48147}}}
\label{fig:fig-compare}
\end{figure*}

Figure \ref{fig:fig3} is an instance of integrating one semantified bioassay in the ORKG DL. This bioassay was semantified on eight semantic statements based on the BAO. Integrating machine actionable graphs of bioassays is essential for the ORKG DL to automatically compute the tabulated comparison surveys of several bioassays as shown in Figure~\ref{fig:fig-compare} in the next section.

\section{Application: Comparisons of Bioassays in ORKG}

Next generation DLs target semantified scholarly knowledge. The ORKG with the semantified bioassays integrated, automatically computes their survey comparisons depending on how many of the machine-actionable assays were selected to be compared by the user. Such tools must be available to scientists to assist them in such massive knowledge ingestion scenarios to quickly grasp the scholarly knowledge highlights fostering faster progress with discoveries.

\end{document}